\def\year{2020}\relax
\documentclass[letterpaper]{article} 
\usepackage{aaai20}  
\usepackage{times}  
\usepackage{helvet} 
\usepackage{courier}  
\usepackage[hyphens]{url}  
\usepackage{graphicx} 
\usepackage{graphicx}  
\usepackage{amsmath}
\usepackage{amssymb}
\usepackage{multirow}
\usepackage{subfigure}
\title{KG-BERT: BERT for Knowledge Graph Completion}
\author{Liang Yao, Chengsheng Mao, Yuan Luo\thanks{Corresponding Author}\\
Northwestern University\\
Chicago IL 60611\\
\{liang.yao, chengsheng.mao, yuan.luo\}@northwestern.edu\\
}
 
\begin{document}

\maketitle

\begin{abstract}
Knowledge graphs are important resources for many artificial intelligence tasks but often suffer from incompleteness. In this work, we propose to use pre-trained language models for knowledge graph completion. We treat triples in knowledge graphs as textual sequences and propose a novel framework named Knowledge Graph Bidirectional Encoder Representations from Transformer (KG-BERT) to model these triples. Our method takes entity and relation descriptions of a triple as input and computes scoring function of the triple with the KG-BERT language model. Experimental results on multiple benchmark knowledge graphs show that our method can achieve state-of-the-art performance in triple classification, link prediction and relation prediction tasks.
\end{abstract}

\section{Introduction}
\noindent Large-scale knowledge graphs (KG) such as FreeBase~\cite{bollacker2008freebase}, YAGO~\cite{suchanek2007yago} and WordNet~\cite{miller1995wordnet} provide effective basis for many important AI tasks such as semantic search, recommendation~\cite{zhang2016collaborative} and question answering~\cite{cui2017kbqa}. A KG is typically a multi-relational graph containing entities as nodes and relations as edges. Each edge is represented as a triplet (\textit{head entity}, relation, \textit{tail entity}) ($(h, r, t)$ for short), indicating the relation between two entities, e.g., (\textit{Steve Jobs}, founded, \textit{Apple Inc.}). Despite their effectiveness, knowledge graphs are still far from being complete. This problem motivates the task of \textit{knowledge graph completion}, which is targeted at assessing the plausibility of triples not present in a knowledge graph.

Much research work has been devoted to knowledge graph completion. A common approach is called knowledge graph embedding which represents entities and relations in triples as real-valued vectors and assess triples' plausibility with these vectors~\cite{wang2017knowledge}. However, most knowledge graph embedding models only use structure information in observed triple facts, which suffer from the sparseness of knowledge graphs. Some recent studies incorporate textual information to enrich knowledge representation~\cite{socher2013reasoning,xie2016representation,xiao2017ssp}, but they learn unique text embedding for the same entity/relation in different triples, which ignore contextual information. For instance, different words in the description of \textit{Steve Jobs} should have distinct importance weights connected to two relations ``founded" and ``isCitizenOf", the relation ``wroteMusicFor" can have two different meanings ``writes lyrics" and ``composes musical compositions" given different entities. On the other hand, syntactic and semantic information in large-scale text data is not fully utilized, as they only employ entity descriptions, relation mentions or word co-occurrence with entities~\cite{wang2016text,xu2017knowledge,an2018accurate}. 


Recently, pre-trained language models such as ELMo~\cite{peters2018deep}, GPT~\cite{radford2018improving}, BERT~\cite{devlin2019bert} and XLNet~\cite{yang2019xlnet} have shown great success in natural language processing (NLP), these models can learn contextualized word embeddings with large amount of free text data and achieve state-of-the-art performance in many language understanding tasks. Among them, BERT is the most prominent one by pre-training the bidirectional Transformer encoder through masked language modeling and next sentence prediction. It can capture rich linguistic knowledge in pre-trained model weights. 

In this study, we propose a novel method for knowledge graph completion using pre-trained language models. Specifically, we first treat entities, relations and triples as textual sequences and turn knowledge graph completion into a sequence classification problem. We then fine-tune BERT model on these sequences for predicting the plausibility of a triple or a relation. The method can achieve strong performance in several KG completion tasks. Our source code is available at \url{https://github.com/yao8839836/kg-bert}. Our contributions are summarized as follows:


\begin{itemize}
    \item We propose a new language modeling method for knowledge graph completion. To the best of our knowledge, this is the first study to model triples' plausibility with a pre-trained contextual language model.
    \item Results on several benchmark datasets show that our method can achieve state-of-the-art results in triple classification, relation prediction and link prediction tasks. 
\end{itemize}

\section{Related Work}
\subsection{Knowledge Graph Embedding}
A literature survey of knowledge graph embedding methods has been conducted by~\cite{wang2017knowledge}. These methods can be classified into translational distance models and semantic matching models based on different scoring functions for a triple $(h,r,t)$. Translational distance models use distance-based scoring functions. They assess the plausibility of a triple $(h,r,t)$ by the distance between the two entity vectors $\mathbf{h}$ and $\mathbf{t}$, typically after a translation performed by the relation vector $\mathbf{r}$. The representative models are TransE~\cite{bordes2013translating} and its extensions including TransH~\cite{wang2014knowledge}. For TransE, the scoring function is defined as the negative translational distance $f(h,r,t) = - || \mathbf{h} + \mathbf{r} - \mathbf{t}||$. Semantic matching models employ similarity-based scoring functions. The representative models are RESCAL~\cite{nickel2011three}, DistMult~\cite{yang2015embedding} and their extensions. For DistMult, the scoring function is defined as a bilinear function $f(h,r,t) = \langle\mathbf{h}, \mathbf{r}, \mathbf{t}\rangle$. Recently, convolutional neural networks also show promising results for knowledge graph completion~\cite{dettmers2018convolutional,SWJ318,schlichtkrull2018modeling}.

The above methods conduct knowledge graph completion using only structural information observed in triples, while different kinds of external information like entity types, logical rules and textual descriptions can be introduced to improve the performance~\cite{wang2017knowledge}. For textual descriptions, \cite{socher2013reasoning} firstly represented entities by averaging the word embeddings contained in their names, where the word embeddings are learned from an external corpus. \cite{wang2014knowledgeb} proposed to jointly embed entities and words into the same vector space by aligning Wikipedia anchors and entity names. \cite{xie2016representation} use convolutional neural networks (CNN) to encode word sequences in entity descriptions. \cite{xiao2017ssp} proposed semantic space projection (SSP) which jointly learns topics and KG embeddings by characterizing the strong correlations between fact triples and textual descriptions. Despite their success, these models learn the same textual representations of entities and relations while words in entity/relation descriptions can have different meanings or importance weights in different triples. 

To address the above problems, \cite{wang2016text} presented a text-enhanced KG embedding model TEKE which can assign different embeddings to a relation in different triples. TEKE utilizes co-occurrences of entities and words in an entity-annotated text corpus. \cite{xu2017knowledge} used an LSTM encoder with attention mechanism to construct contextual text representations given different relations. \cite{an2018accurate} proposed an accurate text-enhanced KG embedding method by exploiting triple specific relation mentions and a mutual attention mechanism between relation mention and entity description. Although these methods can handle the semantic variety of entities and relations in distinct triples, they could not make full use of syntactic and semantic information in large scale free text data, as only entity descriptions, relation mentions and word co-occurrence with entities are utilized. Compared with these methods, our method can learn context-aware text embeddings with rich language information via pre-trained language models.


\subsection{Language Model Pre-training}
Pre-trained language representation models can be divided into two categories: feature-based and fine tuning approaches. Traditional word embedding methods such as Word2Vec~\cite{mikolov2013distributed} and Glove~\cite{pennington2014glove} aimed at adopting feature-based approaches to learn context-independent words vectors. ELMo~\cite{peters2018deep} generalized traditional word embeddings to context-aware word embeddings, where word polysemy can be properly handled. Different from feature-based approaches, fine tuning approaches like GPT~\cite{radford2018improving} and BERT~\cite{devlin2019bert} used the pre-trained model architecture and parameters as a starting point for specific NLP tasks. The pre-trained models capture rich semantic patterns from free text. Recently, pre-trained language models have also been explored in the context of KG. \cite{wang2018dolores} learned contextual embeddings on entity-relation chains (sentences) generated from random walks in KG, then used the embeddings as initialization of KG embeddings models like TransE. \cite{zhang-etal-2019-ernie} incorporated informative entities in KG to enhance BERT language representation. \cite{bosselut-etal-2019-comet} used GPT to generate tail phrase tokens given head phrases and relation types in a common sense knowledge base which does not cleanly fit into a schema comparing two entities with a known relation. The method focuses on generating new entities and relations. Unlike these studies, we use names or descriptions of entities and relations as input and fine-tune BERT to compute plausibility scores of triples.


\section{Method}
\subsection{Bidirectional
Encoder Representations from Transformers (BERT)}

BERT~\cite{devlin2019bert} is a state-of-the-art pre-trained contextual language representation model built on a multi-layer bidirectional Transformer encoder~\cite{vaswani2017attention}. The Transformer encoder is based on self-attention mechanism. There are two steps in BERT framework: \textit{pre-training} and \textit{fine-tuning}. During pre-training, BERT is trained on large-scale unlabeled general domain corpus (3,300M words from BooksCorpus and English Wikipedia) over two self-supervised tasks: masked language modeling and next sentence prediction. In masked language modeling, BERT predicts randomly masked input tokens. In next sentence prediction, BERT predicts whether two input sentences are consecutive. For fine-tuning, BERT is initialized with the pre-trained parameter weights, and all of the parameters are fine-tuned using labeled data from downstream tasks such as sentence pair classification, question answering and sequence labeling.

\subsection{Knowledge Graph BERT (KG-BERT)}

\begin{figure*}[t]
  \centering
  \includegraphics[width = 0.78 \textwidth]{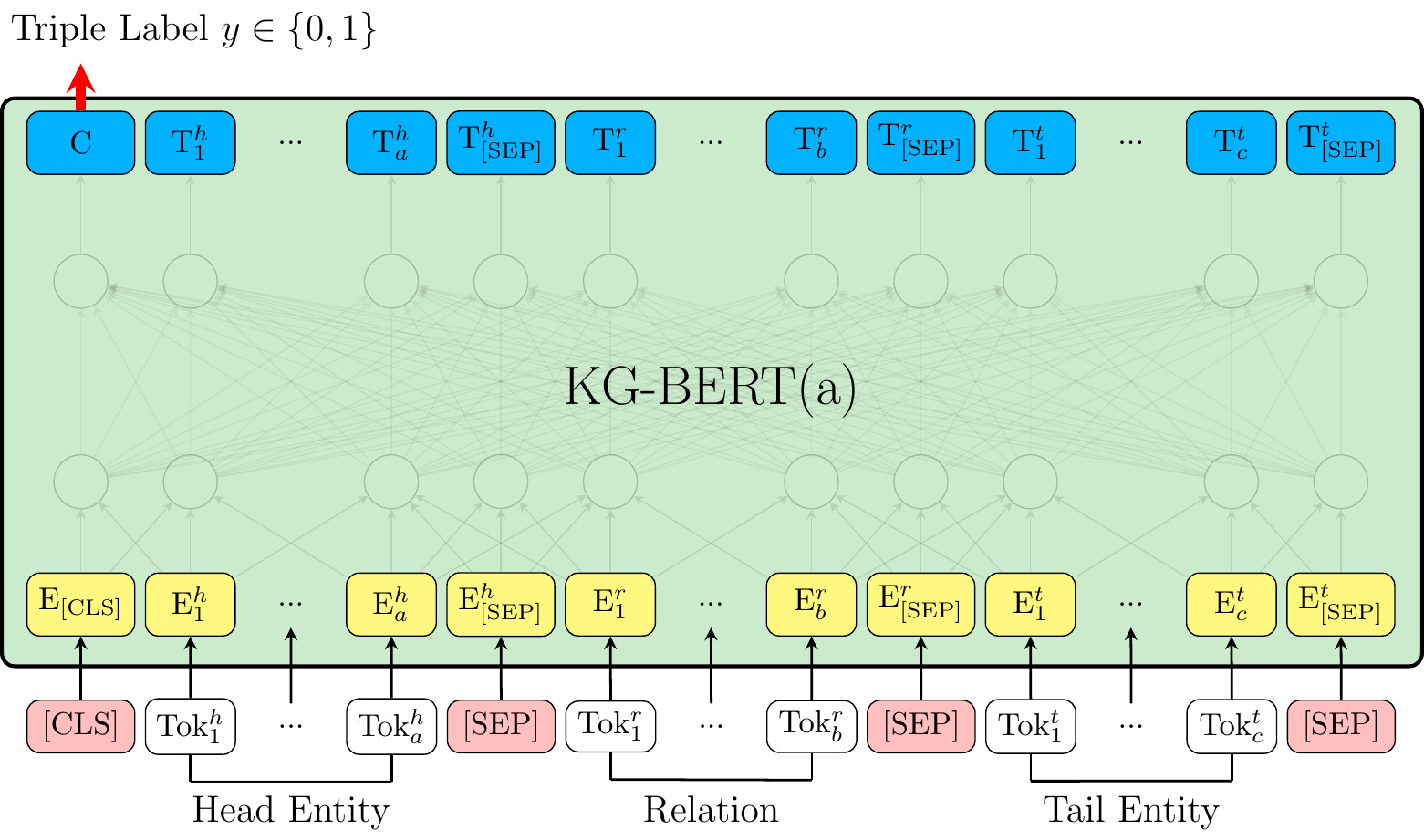}
  \caption{Illustrations of fine-tuning KG-BERT for predicting the plausibility of a triple.}
  \label{fig:framework}
\end{figure*}

\begin{figure}[h]
  \centering
  \includegraphics[width = 0.40 \textwidth]{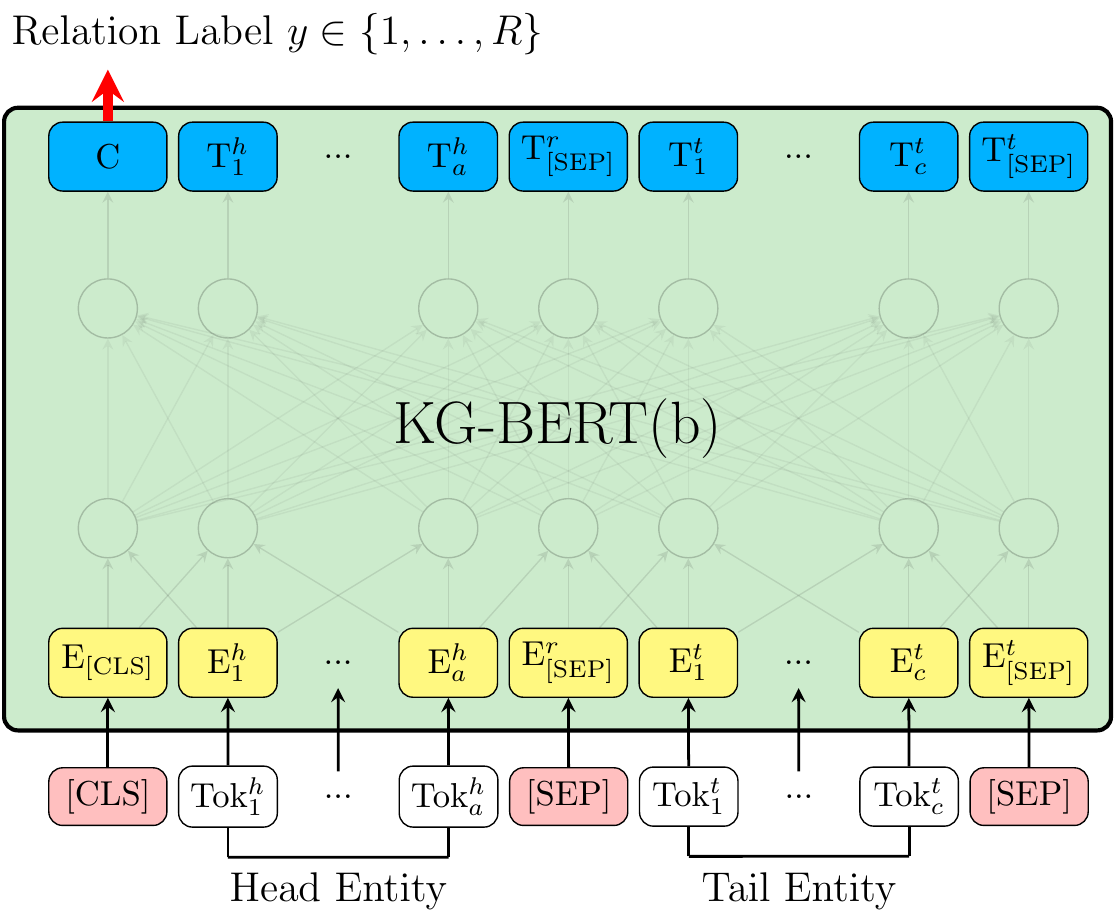}
  \caption{Illustrations of fine-tuning KG-BERT for predicting the relation between two entities.}
  \label{fig:framework}
\end{figure}

To take full advantage of contextual representation with rich language patterns, We fine tune pre-trained BERT for knowledge graph completion. We represent entities and relations as their names or descriptions, then take the name/description word sequences as the input sentence of the BERT model for fine-tuning. As original BERT, a ``sentence" can be an arbitrary span of contiguous text or word sequence, rather than an actual linguistic sentence. To model the plausibility of a triple, we packed the sentences of $(h,r,t)$ as a single sequence.  A “sequence” means the input token sequence to BERT, which may be two entity name/description sentences or three sentences of $(h,r,t)$ packed together. 

The architecture of the KG-BERT for modeling triples is shown in Figure 1. We name this KG-BERT version KG-BERT(a).
The first token of every input sequence is always a special classification token [CLS]. The head entity is represented as a sentence containing tokens Tok$_1^{h}$, ..., Tok$_a^{h}$, e.g., ``\textit{Steven Paul Jobs was an American business magnate, entrepreneur and investor.}" or ``\textit{Steve Jobs}", the relation is represented as a sentence containing tokens Tok$_1^{r}$, ..., Tok$_b^{r}$, e.g., ``founded", the tail entity is represented as a sentence containing tokens Tok$_1^{t}$, ..., Tok$_c^{t}$, e.g., ``\textit{Apple Inc. is an American multinational technology company headquartered in Cupertino, California.}" or ``\textit{Apple Inc.}".  The sentences of entities and relations are separated by a special token [SEP]. For a given token, its input representation is constructed by summing the corresponding token, segment and position embeddings. Different elements separated by [SEP] have different segment embeddings, the tokens in sentences of head and tail entity share the same segment embedding $e_{A}$, while the tokens in relation sentence have a different segment embedding $e_{B}$. Different tokens in the same position $i \in \{1,2,3, \ldots, 512\}$ have a same position embedding. Each input token $i$ has a input representation $E_i$. The token representations are fed into the BERT model architecture which is a multi-layer bidirectional Transformer encoder based on the original implementation described in~\cite{vaswani2017attention}. The final hidden vector of the special [CLS] token and $i$-th input token are denoted as $C \in \mathbb{R}^H$ and $T_i \in \mathbb{R}^H$, where $H$ is the hidden state size in pre-trained BERT. The final hidden state $C$ corresponding to [CLS] is used as the aggregate sequence representation for computing triple scores. The only new parameters introduced during triple classification fine-tuning are classification layer weights $W \in \mathbb{R}^{2 \times H}$. The scoring function for a triple $\tau = (h,r,t)$ is $\mathbf{s_{\tau}} = f(h,r,t) = \text{sigmoid}(CW^T)$, $\mathbf{s}_{\tau} \in \mathbb{R}^2$ is a 2-dimensional real vector with $s_{\tau 0}, s_{\tau 1} \in [0,1]$ and $s_{\tau 0} + s_{\tau 1} = 1$. Given the positive triple set $\mathbb{D}^+$ and a negative triple set $\mathbb{D}^-$ constructed accordingly, we compute a cross-entropy loss with $\mathbf{s}_{\tau}$ and triple labels:

\begin{equation}
\mathcal{L} = -\sum_{\tau \in \mathbb{D}^+ \cup \mathbb{D}^-}{(y_{\tau}\log(s_{\tau 0}) + (1 - y_{\tau})\log(s_{\tau 1}))}
\end{equation}
where $y_{\tau} \in \{0,1\}$ is the label (negative or positive) of that triple. The negative triple set $\mathbb{D}^-$ is simply generated by replacing head entity $h$ or tail entity $t$ in a positive triple $(h,r,t) \in \mathbb{D}^+$ with a random entity $h'$ or $t'$, i.e., 
\begin{equation}
\begin{aligned}
    \mathbb{D}^- = \{(h',r,t)| h' \in \mathbb{E} \land h' \ne h \land (h',r, t) \notin \mathbb{D}^+ \} \\ \cup \{(h,r,t')| t' \in \mathbb{E} \land t' \ne t \land (h,r,t') \notin \mathbb{D}^+\}
\end{aligned}
\end{equation}
where $\mathbb{E}$ is the set of entities. Note that a triple will not be treated as a negative example if it is already in positive set $\mathbb{D}^+$. The pre-trained parameter weights and new weights $W$ can be updated via gradient descent. 
 
The architecture of the KG-BERT for predicting relations is shown in Figure 2. We name this KG-BERT version KG-BERT(b). We only use sentences of the two entities $h$ and $t$ to predict the relation $r$ between them. In our preliminary experiment, we found predicting relations with two entities directly is better than using KG-BERT(a) with relation corruption, i.e., generating negative triples by replacing relation $r$ with a random relation $r'$. As KG-BERT(a), the final hidden state $C$ corresponding to [CLS] is used as the representation of the two entities. The only new parameters introduced in relation prediction fine-tuning are classification layer weights $W' \in \mathbb{R}^{R \times H}$, where $R$ is the number of relations in a KG. The scoring function for a triple $\tau = (h,r,t)$ is $\mathbf{s_{\tau}'} = f(h,r,t) = \text{softmax}(CW'^T)$, $\mathbf{s_{\tau}'} \in \mathbb{R}^R$ is a $R$-dimensional real vector with $s'_{\tau i} \in [0,1]$ and $\sum_{i}^R s_{\tau i}' = 1$. We compute the following cross-entropy loss with $\mathbf{s'}_{\tau}$ and relation labels:
\begin{equation}
\mathcal{L'} = -\sum_{\tau \in \mathbb{D}^+ }\sum_{i=1}^R{y'_{\tau i}\log(s'_{\tau i})}
\end{equation}
where $\tau$ is an observed positive triple, $y'_{\tau i}$ is the relation indicator for the triple $\tau$, $y'_{\tau i} = 1$ when $r=i$ and $y'_{\tau i} = 0$ when $r \ne i$.
\section{Experiments}

In this section we evaluate our KG-BERT on three experimental tasks. Specifically we want to determine:
\begin{itemize}
    \item Can our model judge whether an unseen triple fact $(h,r,t)$ is true or not?
    \item Can our model predict an entity given another entity and a specific relation?
    \item Can our model predict relations given two entities?
\end{itemize}

\paragraph{Datasets.}
We ran our experiments on six widely used benchmark KG datasets: WN11~\cite{socher2013reasoning}, FB13~\cite{socher2013reasoning}, FB15K~\cite{bordes2013translating}, WN18RR, FB15k-237 and UMLS~\cite{dettmers2018convolutional}. WN11 and WN18RR are two subsets of WordNet, FB15K and FB15k-237 are two subsets of Freebase. WordNet is a large lexical KG of English where each entity as a synset which is consisting of several words and corresponds to a distinct word sense. Freebase is a large knowledge graph of general world facts. UMLS is a medical semantic network containing semantic types (entities) and semantic relations. The test sets of WN11 and FB13 contain positive and negative triplets which can be used for triple classification. The test set of WN18RR, FB15K, FB15k-237 and UMLS only contain correct triples, we perform link (entity) prediction and relation prediction on these datasets. Table 1 provides statistics of all datasets we used. 


For WN18RR, we use synsets definitions as entity sentences. For WN11, FB15K and UMLS, we use entity names as input sentences. For FB13, we use entity descriptions in Wikipedia as input sentences. For FB15k-237, we used entity descriptions made by ~\cite{xie2016representation}. For all datasets, we use relation names as relation sentences.

    {\small
    \begin{table}[t]\footnotesize
    \centering
    \renewcommand{\arraystretch}{1.2}

    \begin{tabular}{c|ccccccc}
    \hline
    \bf{Dataset}& \bf{\# Ent}	& \bf{\# Rel}& \bf{\# Train}& \bf{\# Dev} & \bf{\# Test} \\
    \hline
    WN11 & 38,696 & 11 & 112,581 & 2,609  & 10,544\\ 
    FB13 & 75,043  & 13 & 316,232 & 5,908 & 23,733 \\
    WN18RR& 40,943 & 11 & 86,835 & 3,034 & 3,134\\
    FB15K& 14,951 & 1,345 & 483,142 & 50,000 & 59,071 \\
    FB15k-237& 14,541 & 237 & 272,115 & 17,535 & 20,466\\
    UMLS & 135 & 46 & 5,216 & 652 & 661\\
    \hline
    \end{tabular}
    \caption{Summary statistics of datasets.}
    \label{tab:statistics}
    \end{table}
    }
    

\paragraph{Baselines.}
We compare our KG-BERT with multiple state-of-the-art KG embedding methods as follows: TransE and its extensions TransH~\cite{wang2014knowledge}, TransD~\cite{ji2015knowledge}, TransR~\cite{lin2015learning}, TransG~\cite{xiao2016transg}, TranSparse~\cite{ji2016knowledge} and PTransE~\cite{lin2015modeling}, DistMult and its extension DistMult-HRS~\cite{zhang2018knowledge} which only used structural information in KG. The neural tensor network NTN~\cite{socher2013reasoning} and its simplified version ProjE~\cite{shi2017proje}. CNN models: ConvKB~\cite{SWJ318}, ConvE~\cite{dettmers2018convolutional} and R-GCN~\cite{schlichtkrull2018modeling}. KG embeddings with textual information: TEKE~\cite{wang2016text}, DKRL~\cite{xie2016representation}, SSP~\cite{xiao2017ssp}, AATE~\cite{an2018accurate}. KG embeddings with entity hierarchical types: TKRL~\cite{xie2016representationijcai}. Contextualized KG embeddings: DOLORES~\cite{wang2018dolores}. Complex-valued KG embeddings ComplEx~\cite{trouillon2016complex} and RotatE~\cite{sun2019rotate}. Adversarial learning framework: KBGAN~\cite{cai2018kbgan}. 

\paragraph{Settings.}
We choose pre-trained BERT-Base model with 12 layers, 12 self-attention heads and $H=768$ as the initialization of KG-BERT, then fine tune KG-BERT with Adam implemented in BERT. In our preliminary experiment, we found BERT-Base model can achieve better results than BERT-Large in general, and BERT-Base is simpler and less sensitive to hyper-parameter choices. Following original BERT, we set the following hyper-parameters in KG-BERT fine-tuning: batch size: 32, learning rate: 5e-5, dropout rate: 0.1. We also tried other values of these hyper-parameters in~\cite{devlin2019bert} but didn't find much difference. We tuned number of epochs for different tasks: 3 for triple classification, 5 for link (entity) prediction and 20 for relation prediction. We found more epochs can lead to better results in relation prediction but not in other two tasks. For triple classification training, we sample 1 negative triple for a positive triple which can ensure class balance in binary classification. For link (entity) prediction training, we sample 5 negative triples for a positive triple, we tried 1, 3, 5 and 10 and found 5 is the best.

    {\small
    \begin{table}[h]\scriptsize
    \centering
    \renewcommand{\arraystretch}{1.1}

    \begin{tabular}{l|cc|c}
    \hline
    \bf{Method}& WN11& FB13& Avg. \\
    \hline
    NTN~\cite{socher2013reasoning}& 86.2 & 90.0 & 88.1\\
    TransE~\cite{wang2014knowledge} & 75.9 & 81.5 & 78.7 \\   
    TransH~\cite{wang2014knowledge} & 78.8 & 83.3 & 81.1 \\
    TransR~\cite{lin2015learning} & 85.9 & 82.5 & 84.2 \\
    TransD~\cite{ji2015knowledge} & 86.4  & 89.1 & 87.8 \\ 
    TEKE~\cite{wang2016text} & 86.1 & 84.2 & 85.2 \\ 
    TransG~\cite{xiao2016transg} & 87.4  & 87.3  & 87.4  \\
    TranSparse-S~\cite{ji2016knowledge} & 86.4& 88.2& 87.3\\
    DistMult~\cite{zhang2018knowledge} & 87.1& 86.2 &86.7 \\
    DistMult-HRS~\cite{zhang2018knowledge} & 88.9& 89.0 &89.0 \\
    AATE~\cite{an2018accurate} & 88.0  & 87.2  & 87.6 \\    
    ConvKB~\cite{SWJ318} & 87.6  & 88.8 & 88.2   \\
    DOLORES~\cite{wang2018dolores} & 87.5 & 89.3  & 88.4   \\
    KG-BERT(a) & \textbf{93.5} & \textbf{90.4} & \textbf{91.9} \\
    \hline
    \end{tabular}
    \caption{Triple classification accuracy (in percentage) for different embedding methods. The baseline results are obtained from corresponding papers.}
    \label{tab:statistics}
    \end{table}
    }

    {\small
    \begin{table*}[h]
    \centering
    \renewcommand{\arraystretch}{1.2}

    \begin{tabular}{l|cc|cc|cc}
    
    \hline
    \multirow{2}*{Method} &\multicolumn{2}{c|}{WN18RR} & \multicolumn{2}{c|}{FB15k-237} & \multicolumn{2}{c}{UMLS}\\
    \cline{2-7}
    & MR & Hits@10  & MR & Hits@10 & MR & Hits@10\\
    \hline
    TransE (our results) & 2365 & 50.5 &223 & 47.4 &1.84&98.9\\ 
    TransH (our results)& 2524 & 50.3 & 255& 48.6 &1.80& \textbf{99.5}\\ 
    TransR (our results)& 3166 & 50.7 &237 &51.1  &1.81& 99.4\\ 
    TransD (our results)& 2768 & 50.7 & 246 & 48.4 &1.71 & 99.3 \\ 
    DistMult (our results)  & 3704& 47.7 &411 & 41.9&5.52 & 84.6\\ 
    ComplEx (our results)&  3921 & 48.3 & 508& 43.4 &2.59 & 96.7\\ 
    ConvE~\cite{dettmers2018convolutional} & 5277 & 48 & 246&  49.1& -- & --\\
    ConvKB~\cite{SWJ318} & 2554 & 52.5 & 257& 51.7 &--&-- \\
    R-GCN~\cite{schlichtkrull2018modeling} & -- & -- & --& 41.7 &--& --\\
    KBGAN~\cite{cai2018kbgan} & --& 48.1 & --& 45.8& --&--\\
    RotatE~\cite{sun2019rotate} &3340& \textbf{57.1} & 177& \textbf{53.3}& --&--\\    
    KG-BERT(a) & \textbf{97} &52.4 & \textbf{153} &42.0 &\textbf{1.47}&  99.0\\
    \hline
    \end{tabular}
    \caption{Link prediction results on WN18RR, FB15k-237 and UMLS datasets. The baseline models denoted (our results) are implemented using OpenKE toolkit~\cite{han2018openke}, other baseline results are taken from the original papers.}
    \label{tab:statistics}
    \end{table*}
    }

    {\small
    \begin{table}[h]\scriptsize
    \centering
    \renewcommand{\arraystretch}{1.1}
    \begin{tabular}{l|cc}
    \hline
    \bf{Method}& Mean Rank & Hits@1 \\
    \hline
    TransE~\cite{lin2015modeling} & 2.5 & 84.3  \\ 
    TransR~\cite{xie2016representationijcai} &2.1 & 91.6\\
    DKRL (CNN)~\cite{xie2016representation}& 2.5  & 89.0    \\
    DKRL (CNN) + TransE~\cite{xie2016representation}& 2.0  & 90.8    \\
    DKRL (CBOW)~\cite{xie2016representation} & 2.5  & 82.7    \\
    TKRL (RHE)~\cite{xie2016representationijcai}  & 1.7  & 92.8    \\
    TKRL (RHE)~\cite{xie2016representationijcai}  & 1.8  & 92.5    \\
    PTransE (ADD, len-2 path)~\cite{lin2015modeling} & \textbf{1.2}& 93.6 \\
    PTransE (RNN, len-2 path)~\cite{lin2015modeling}  & 1.4  & 93.2 \\    
    PTransE (ADD, len-3 path)~\cite{lin2015modeling} &  1.4  &  94.0 \\
    SSP~\cite{xiao2017ssp} &\textbf{1.2}&--\\
    ProjE (pointwise)~\cite{shi2017proje} & 1.3 &  95.6   \\    
    ProjE (listwise)~\cite{shi2017proje} & \textbf{1.2} &  95.7   \\
    ProjE (wlistwise)~\cite{shi2017proje} & \textbf{1.2} &  95.6   \\   
    KG-BERT (b) & \textbf{1.2} & \textbf{96.0}  \\
    \hline
    \end{tabular}
    \caption{Relation prediction results on FB15K dataset. The baseline results are obtained from corresponding papers. }
    \label{tab:statistics}
    \end{table}
    }

\paragraph{Triple Classification.}
Triple classification aims to judge whether a given triple $(h, r, t)$ is correct or not. Table 2 presents triple classification accuracy of different methods on WN11 and FB13. We can see that KG-BERT(a) clearly outperforms all baselines by a large margin, which shows the effectiveness of our method. We ran our models 10 times and found the standard deviations are less than 0.2, and the improvements are significant ($p <0.01$). To our knowledge, KG-BERT(a) achieves the best results so far. For more in-depth performance analysis, we note that TransE could not achieve high accuracy scores because it could not deal with 1-to-N, N-to-1, and N-to-N relations. TransH, TransR, TransD, TranSparse and TransG outperform TransE by introducing relation specific parameters. DistMult performs relatively well, and can also be improved by hierarchical relation structure information used in DistMult-HRS. ConvKB shows decent results, which suggests that CNN models can capture global interactions among the entity and relation embeddings. DOLORES further improves ConvKB by incorporating contextual information in entity-relation random walk chains. NTN also achieves competitive performances especially on FB13, which means it's an expressive model, and representing entities with word embeddings is helpful. Other text-enhanced KG embeddings TEKE and AATE outperform their base models like TransE and TransH, which demonstrates the benefit of external text data. However, their improvements are still limited due to less utilization of rich language patterns. The improvement of KG-BERT(a) over baselines on WN11 is larger than FB13, because WordNet is a linguistic knowledge graph which is closer to linguistic patterns contained in pre-trained language models.

Figure 3 reports triple classification accuracy with 5$\%$, 10$\%$, 15$\%$, 20$\%$ and 30$\%$ of original WN11 and FB13 training triples. We note that KG-BERT(a) can achieve higher test accuracy with limited training triples. For instance, KG-BERT(a) achieves a test accuracy of 88.1$\%$ on FB13 with only $5\%$ training triples and a test accuracy of 87.0$\%$ on WN11 with only $10\%$ training triples which are higher than some baseline models (including text-enhanced models) with even the full training triples. These encouraging results suggest that KG-BERT(a) can fully utilize rich linguistic patterns in large external text data to overcome the sparseness of knowledge graphs.

The main reasons why KG-BERT(a) performs well are four fold: 1) The input sequence contains both entity and relation word sequences; 2) The triple classification task is very similar to next sentence prediction task in BERT pre-training which captures relationship between two sentences in large free text, thus the pre-trained BERT weights are well positioned for the inference of relationship among different elements in a triple; 3) The token hidden vectors are contextual embeddings. The same token can have different hidden vectors in different triples, thus contextual information is explicitly used. 4) The self-attention mechanism can discover the most important words connected to the triple fact.

\begin{figure}[t]
\centering
\subfigure[WN11]{
\label{fig:proportion:a} %
\includegraphics[height = 28 mm]{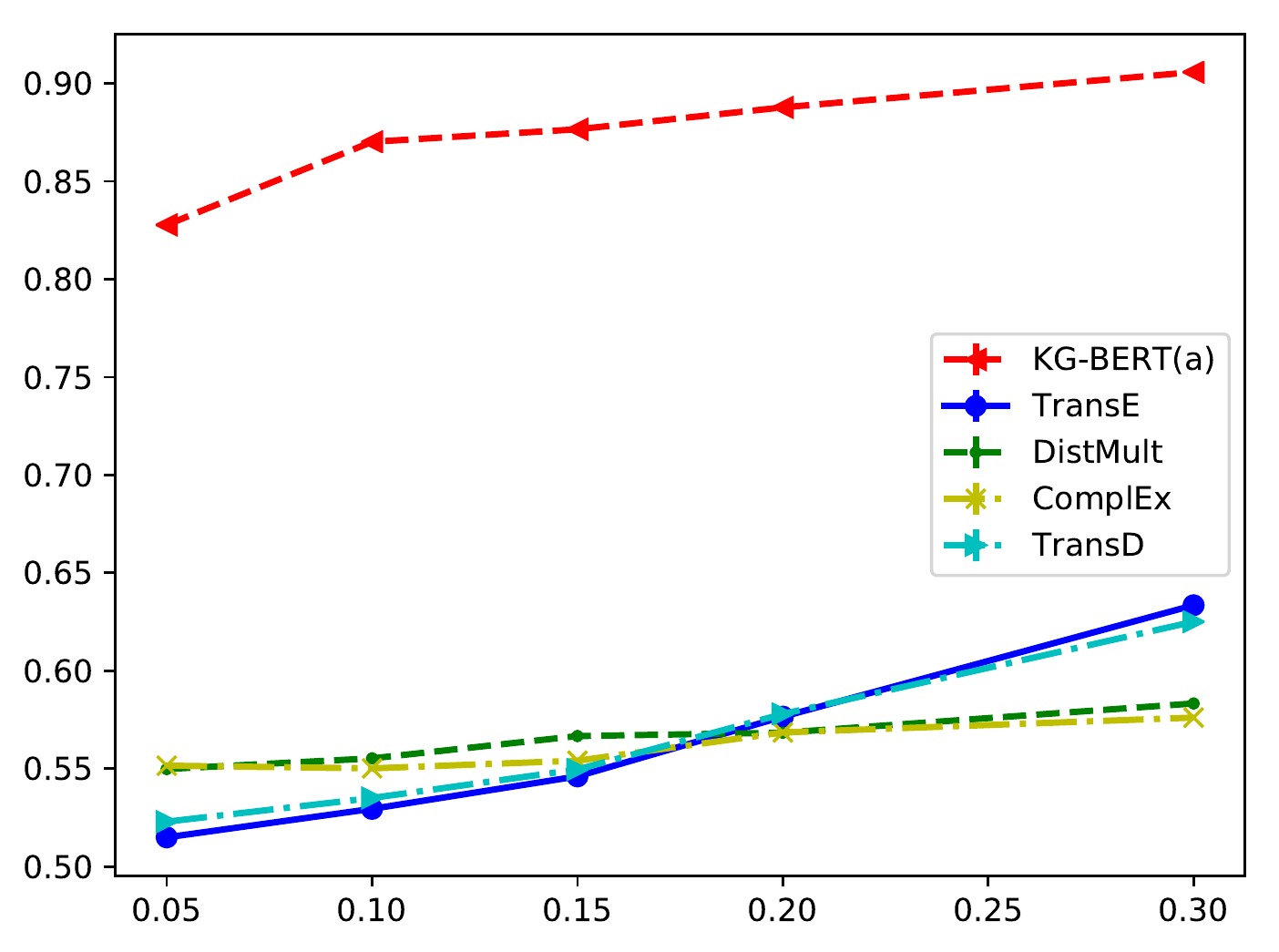}}
\subfigure[FB13]{
\label{fig:proportion:b} %
\includegraphics[height = 28 mm]{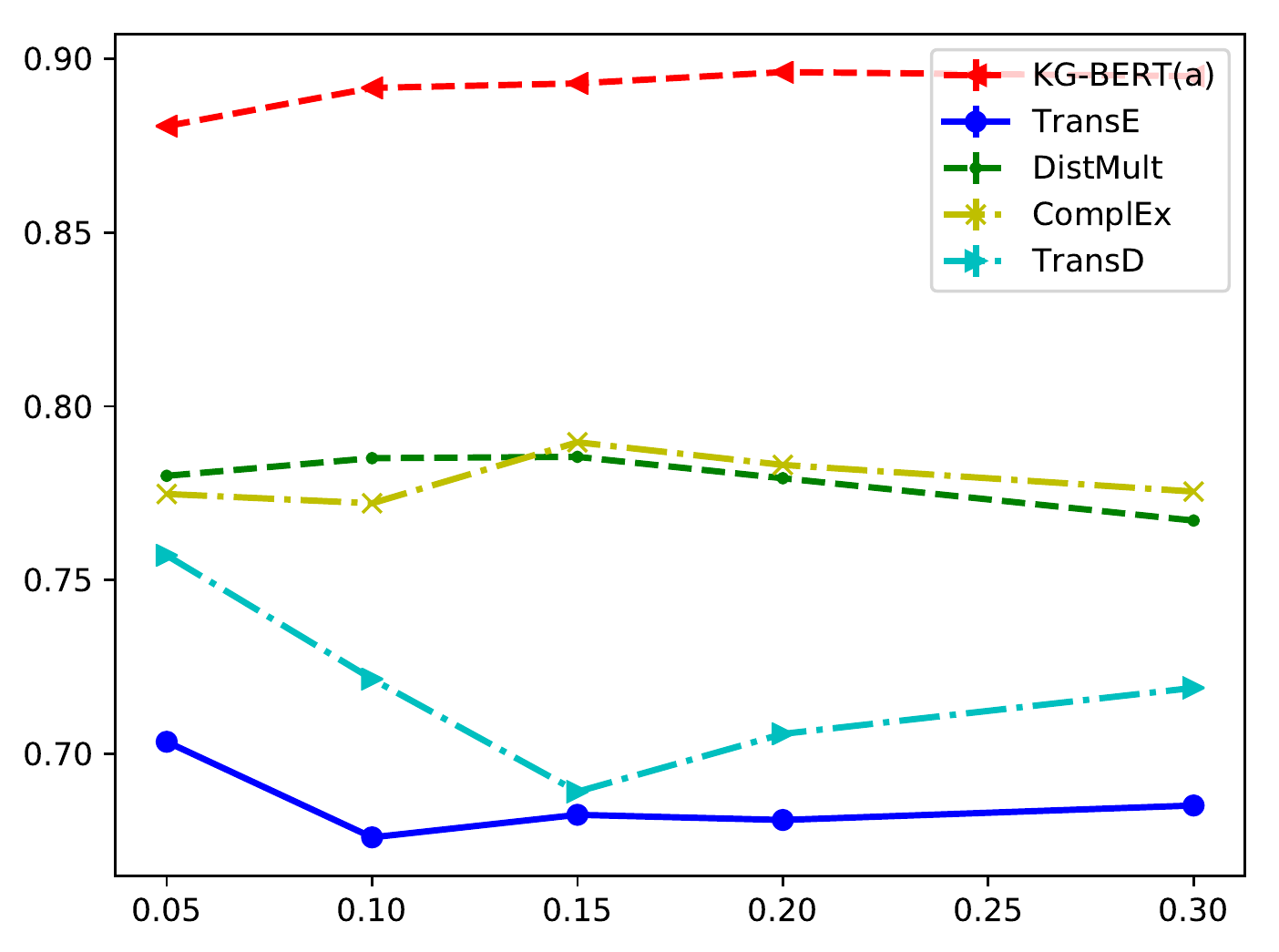}}
\caption{Test accuracy of triple classification by varying training data proportions.}
\label{fig:proportion}
\end{figure}

\paragraph{Link Prediction.}
The link (entity) prediction task predicts the head entity $h$ given $(?, r, t)$ or predicts the tail entity $t$ given $(h, r, ?)$ where $?$ means the missing element. The results are evaluated using a ranking produced by the scoring function $f(h, r, t)$ ($s_{\tau 0}$ in our method) on test triples. Each correct test triple $(h, r, t)$ is corrupted by replacing either its head or tail entity with every entity $e \in \mathbb{E}$, then these candidates are ranked in descending order of their plausibility score. We report two common metrics, Mean Rank (MR) of correct entities and Hits@10 which means the proportion of correct entities in top 10. A lower MR is better while a higher Hits@10 is better. Following~\cite{nguyen2018novel}, we only report results under the \textit{filtered} setting~\cite{bordes2013translating} which removes all corrupted triples appeared in training, development, and test set before getting the ranking lists. 

Table 3 shows link prediction performance of various models. We test some classical baseline models with OpenKE toolkit~\cite{han2018openke}\footnote{https://github.com/thunlp/OpenKE}, other results are taken from the original papers. We can observe that: 1) KG-BERT(a) can achieve lower MR than baseline models, and it achieves the lowest mean ranks on WN18RR and FB15k-237 to our knowledge. 2) The Hits@10 scores of KG-BERT(a) is lower than some state-of-the-art methods. KG-BERT(a) can avoid very high ranks with semantic relatedness of entity and relation sentences, but the KG structure information is not explicitly modeled, thus it could not rank some neighbor entities of a given entity in top 10. CNN models ConvE and ConvKB perform better compared to the graph convolutional network R-GCN. ComplEx could not perform well on WN18RR and FB15k-237, but can be improved using adversarial negative sampling in KBGAN and RotatE.

 \paragraph{Relation Prediction.}
This task predicts relations between two given entities, i.e., $(h,?,t)$. The procedure is similar to link prediction while we rank the candidates with the relation scores $\mathbf{s_{\tau}'}$. We evaluate the relation ranking using Mean Rank (MR) and Hits@1 with \textit{filtered} setting.

Table 4 reports relation prediction results on FB15K. We note that KG-BERT(b) also shows promising results and achieves the highest Hits@1 so far. The KG-BERT(b) is analogous to sentence pair classification in BERT fine-tuning and can also benefit from BERT pre-training. Text-enhanced models DKRL and SSP can also outperform structure only methods TransE and TransH. TKRL and PTransE work well with hierarchical entity categories and extended path information. ProjE achieves very competitive results by treating KG completion as a ranking problem and optimizing ranking score vectors.

\begin{figure}[h]
  \centering
  \includegraphics[width = 0.40 \textwidth]{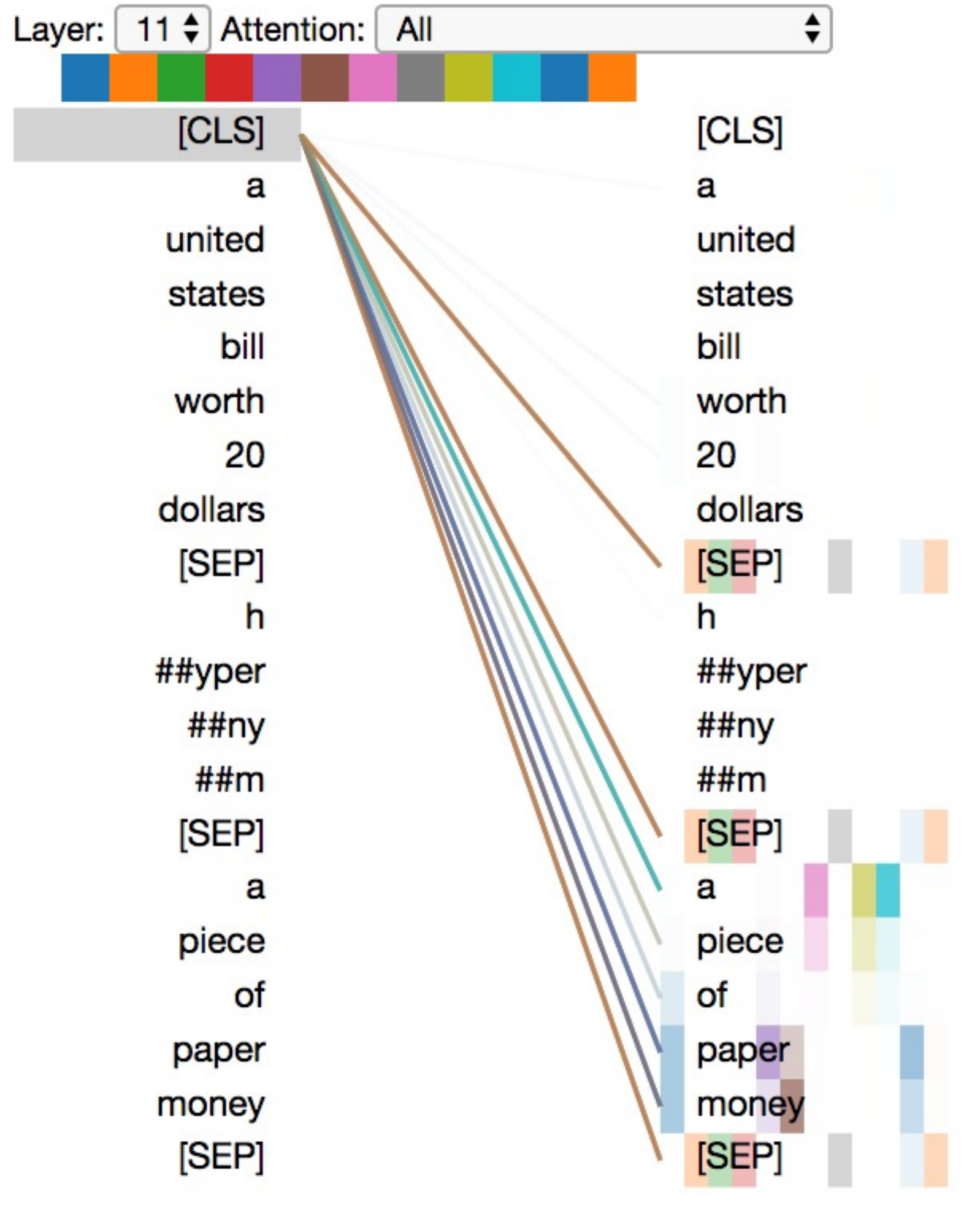}
  \caption{Illustrations of attention patterns of KG-BERT(a). A positive training triple (\textit{\underline{~~~~}twenty\underline{~~}dollar\underline{~~}bill\underline{~~}NN\underline{~~}1}, ~\underline{~~}hypernym, \textit{\underline{~~~~}note\underline{~~}NN\underline{~~}6}) from WN18RR is used as the example.  Different colors mean different attention heads. Transparencies of colors reflect the attention scores. We show the attention weights between [CLS] and other tokens in layer 11 of the Transformer model.}
  \label{fig:framework}
\end{figure}

\begin{figure}[h]
  \centering
  \includegraphics[width = 0.40 \textwidth]{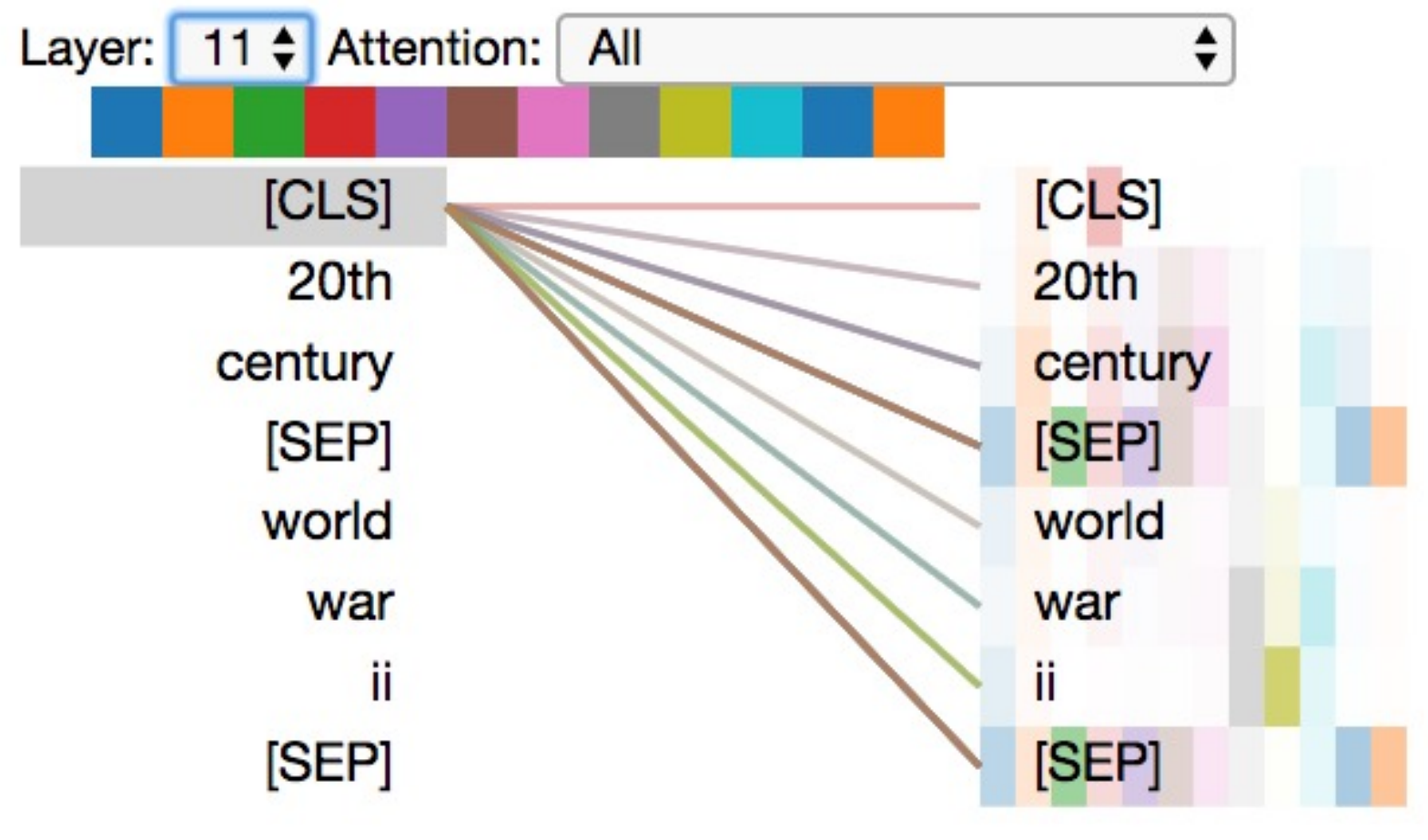}
  \caption{Illustrations of attention patterns of KG-BERT(b). The example is taken from FB15K. Two entities \textit{20th century} and \textit{World War II} are used as input, the relation label is /time/event/includes\underline{~~}event. }
  \label{fig:framework}
\end{figure}

\paragraph{Attention Visualization.}
We show attention patterns of KG-BERT in Figure 4 and Figure 5. We use the visualization tool released by~\cite{vig2019transformervis}\footnote{\url{https://github.com/jessevig/bertviz}}. Figure 4 depicts the attention patterns of KG-BERT(a). A positive training triple (\textit{\underline{~~~~}twenty\underline{~~}dollar\underline{~~}bill\underline{~~}NN\underline{~~}1}, ~\underline{~~}hypernym, \textit{\underline{~~~~}note\underline{~~}NN\underline{~~}6}) from WN18RR is taken as the example. The entity descriptions ``a United States bill worth 20 dollars" and ``a piece of paper money" as well as the relation name ``hypernym" are used as the input sequence. We observe that some important words such as ``paper" and ``money" have higher attention scores connected to the label token [CLS], while some less related words like ``united" and ``states" obtain less attentions. On the other hand, we can see that different attention heads focus on different tokens. [SEP] is highlighted by the same six attention heads, ``a" and ``piece" are highlighted by the three same attention heads, while ``paper" and ``money" are highlighted by other four attention heads. As mentioned in~\cite{vaswani2017attention}, multi-head attention allows KG-BERT to jointly attend to information from different representation subspaces at different positions, different attention heads are concatenated to compute the final attention values. 
Figure 5 illustrates attention patterns of KG-BERT(b). The triple (\textit{20th century}, /time/event/includes\underline{~~}event, \textit{World War II}) from FB15K is taken as input. We can see similar attention patterns as in KG-BERT(a), six attention heads attend to ``century" in head entity, while other three attention heads focus on ``war" and ``ii" in tail entity. Multi-head attention can attend to different aspects of two entities in a triple.

\paragraph{Discussions.}
From experimental results, we note that KG-BERT can achieve strong performance in three KG completion tasks. However, a major limitation is that BERT model is expensive, which makes the link prediction evaluation very time consuming, link prediction evaluation needs to replace head or tail entity with almost all entities, and all corrupted triple sequences are fed into the 12 layer Transformer model. Possible solutions are introducing 1-N scoring models like ConvE or using lightweight language models.

 \section{Conclusion and Future Work}
 In this work, we propose a novel knowledge graph completion method termed Knowledge Graph BERT (KG-BERT). We represent entities and relations as their name/description textual sequences, and turn knowledge graph completion problem into a sequence classification problem. KG-BERT can make use of rich language information in large amount free text and highlight most important words connected to a triple. The proposed method demonstrates promising results by outperforming state-of-the-art results on multiple benchmark KG datasets.
 
 Some future directions include improving the results by jointly modeling textual information with KG structures, or utilizing pre-trained models with more text data like XLNet. And applying our KG-BERT as a knowledge-enhanced language model to language understanding tasks is an interesting future work we are going to explore.


\bibliographystyle{aaai}
\bibliography{kg_bert}

\begin{thebibliography}{}

\bibitem[\protect\citeauthoryear{An \bgroup et al\mbox.\egroup
  }{2018}]{an2018accurate}
An, B.; Chen, B.; Han, X.; and Sun, L.
\newblock 2018.
\newblock Accurate text-enhanced knowledge graph representation learning.
\newblock In {\em NAACL},  745--755.

\bibitem[\protect\citeauthoryear{Bollacker \bgroup et al\mbox.\egroup
  }{2008}]{bollacker2008freebase}
Bollacker, K.; Evans, C.; Paritosh, P.; Sturge, T.; and Taylor, J.
\newblock 2008.
\newblock Freebase: a collaboratively created graph database for structuring
  human knowledge.
\newblock In {\em SIGMOD},  1247--1250.

\bibitem[\protect\citeauthoryear{Bordes \bgroup et al\mbox.\egroup
  }{2013}]{bordes2013translating}
Bordes, A.; Usunier, N.; Garcia-Duran, A.; Weston, J.; and Yakhnenko, O.
\newblock 2013.
\newblock Translating embeddings for modeling multi-relational data.
\newblock In {\em NIPS},  2787--2795.

\bibitem[\protect\citeauthoryear{Bosselut \bgroup et al\mbox.\egroup
  }{2019}]{bosselut-etal-2019-comet}
Bosselut, A.; Rashkin, H.; Sap, M.; Malaviya, C.; Celikyilmaz, A.; and Choi, Y.
\newblock 2019.
\newblock {COMET}: Commonsense transformers for automatic knowledge graph
  construction.
\newblock In {\em ACL},  4762--4779.

\bibitem[\protect\citeauthoryear{Cai and Wang}{2018}]{cai2018kbgan}
Cai, L., and Wang, W.~Y.
\newblock 2018.
\newblock {KBGAN}: Adversarial learning for knowledge graph embeddings.
\newblock In {\em NAACL},  1470--1480.

\bibitem[\protect\citeauthoryear{Cui \bgroup et al\mbox.\egroup
  }{2017}]{cui2017kbqa}
Cui, W.; Xiao, Y.; Wang, H.; Song, Y.; Hwang, S.-w.; and Wang, W.
\newblock 2017.
\newblock {KBQA}: learning question answering over qa corpora and knowledge
  bases.
\newblock {\em Proceedings of the VLDB Endowment} 10(5):565--576.

\bibitem[\protect\citeauthoryear{Dettmers \bgroup et al\mbox.\egroup
  }{2018}]{dettmers2018convolutional}
Dettmers, T.; Minervini, P.; Stenetorp, P.; and Riedel, S.
\newblock 2018.
\newblock Convolutional 2d knowledge graph embeddings.
\newblock In {\em AAAI},  1811--1818.

\bibitem[\protect\citeauthoryear{Devlin \bgroup et al\mbox.\egroup
  }{2019}]{devlin2019bert}
Devlin, J.; Chang, M.-W.; Lee, K.; and Toutanova, K.
\newblock 2019.
\newblock Bert: Pre-training of deep bidirectional transformers for language
  understanding.
\newblock In {\em NAACL},  4171--4186.

\bibitem[\protect\citeauthoryear{Han \bgroup et al\mbox.\egroup
  }{2018}]{han2018openke}
Han, X.; Cao, S.; Lv, X.; Lin, Y.; Liu, Z.; Sun, M.; and Li, J.
\newblock 2018.
\newblock {OpenKE}: An open toolkit for knowledge embedding.
\newblock In {\em EMNLP},  139--144.

\bibitem[\protect\citeauthoryear{Ji \bgroup et al\mbox.\egroup
  }{2015}]{ji2015knowledge}
Ji, G.; He, S.; Xu, L.; Liu, K.; and Zhao, J.
\newblock 2015.
\newblock Knowledge graph embedding via dynamic mapping matrix.
\newblock In {\em ACL},  687--696.

\bibitem[\protect\citeauthoryear{Ji \bgroup et al\mbox.\egroup
  }{2016}]{ji2016knowledge}
Ji, G.; Liu, K.; He, S.; and Zhao, J.
\newblock 2016.
\newblock Knowledge graph completion with adaptive sparse transfer matrix.
\newblock In {\em AAAI}.

\bibitem[\protect\citeauthoryear{Lin \bgroup et al\mbox.\egroup
  }{2015a}]{lin2015modeling}
Lin, Y.; Liu, Z.; Luan, H.; Sun, M.; Rao, S.; and Liu, S.
\newblock 2015a.
\newblock Modeling relation paths for representation learning of knowledge
  bases.
\newblock In {\em EMNLP},  705--714.

\bibitem[\protect\citeauthoryear{Lin \bgroup et al\mbox.\egroup
  }{2015b}]{lin2015learning}
Lin, Y.; Liu, Z.; Sun, M.; Liu, Y.; and Zhu, X.
\newblock 2015b.
\newblock Learning entity and relation embeddings for knowledge graph
  completion.
\newblock In {\em AAAI}.

\bibitem[\protect\citeauthoryear{Mikolov \bgroup et al\mbox.\egroup
  }{2013}]{mikolov2013distributed}
Mikolov, T.; Sutskever, I.; Chen, K.; Corrado, G.~S.; and Dean, J.
\newblock 2013.
\newblock Distributed representations of words and phrases and their
  compositionality.
\newblock In {\em NIPS},  3111--3119.

\bibitem[\protect\citeauthoryear{Miller}{1995}]{miller1995wordnet}
Miller, G.~A.
\newblock 1995.
\newblock Wordnet: a lexical database for english.
\newblock {\em Communications of the ACM} 38(11):39--41.

\bibitem[\protect\citeauthoryear{Nguyen \bgroup et al\mbox.\egroup
  }{2018a}]{SWJ318}
Nguyen, D.~Q.; Nguyen, D.~Q.; Nguyen, T.~D.; and Phung, D.
\newblock 2018a.
\newblock A convolutional neural network-based model for knowledge base
  completion and its application to search personalization.
\newblock {\em Semantic Web}.

\bibitem[\protect\citeauthoryear{Nguyen \bgroup et al\mbox.\egroup
  }{2018b}]{nguyen2018novel}
Nguyen, D.~Q.; Nguyen, T.~D.; Nguyen, D.~Q.; and Phung, D.
\newblock 2018b.
\newblock A novel embedding model for knowledge base completion based on
  convolutional neural network.
\newblock In {\em NAACL},  327--333.

\bibitem[\protect\citeauthoryear{Nickel, Tresp, and
  Kriegel}{2011}]{nickel2011three}
Nickel, M.; Tresp, V.; and Kriegel, H.-P.
\newblock 2011.
\newblock A three-way model for collective learning on multi-relational data.
\newblock In {\em ICML},  809--816.

\bibitem[\protect\citeauthoryear{Pennington, Socher, and
  Manning}{2014}]{pennington2014glove}
Pennington, J.; Socher, R.; and Manning, C.
\newblock 2014.
\newblock Glove: Global vectors for word representation.
\newblock In {\em EMNLP}.

\bibitem[\protect\citeauthoryear{Peters \bgroup et al\mbox.\egroup
  }{2018}]{peters2018deep}
Peters, M.~E.; Neumann, M.; Iyyer, M.; Gardner, M.; Clark, C.; Lee, K.; and
  Zettlemoyer, L.
\newblock 2018.
\newblock Deep contextualized word representations.
\newblock In {\em NAACL},  2227--2237.

\bibitem[\protect\citeauthoryear{Radford \bgroup et al\mbox.\egroup
  }{2018}]{radford2018improving}
Radford, A.; Narasimhan, K.; Salimans, T.; and Sutskever, I.
\newblock 2018.
\newblock Improving language understanding by generative pre-training.

\bibitem[\protect\citeauthoryear{Schlichtkrull \bgroup et al\mbox.\egroup
  }{2018}]{schlichtkrull2018modeling}
Schlichtkrull, M.; Kipf, T.~N.; Bloem, P.; Van Den~Berg, R.; Titov, I.; and
  Welling, M.
\newblock 2018.
\newblock Modeling relational data with graph convolutional networks.
\newblock In {\em ESWC},  593--607.

\bibitem[\protect\citeauthoryear{Shi and Weninger}{2017}]{shi2017proje}
Shi, B., and Weninger, T.
\newblock 2017.
\newblock {ProjE}: Embedding projection for knowledge graph completion.
\newblock In {\em AAAI}.

\bibitem[\protect\citeauthoryear{Socher \bgroup et al\mbox.\egroup
  }{2013}]{socher2013reasoning}
Socher, R.; Chen, D.; Manning, C.~D.; and Ng, A.
\newblock 2013.
\newblock Reasoning with neural tensor networks for knowledge base completion.
\newblock In {\em NIPS},  926--934.

\bibitem[\protect\citeauthoryear{Suchanek, Kasneci, and
  Weikum}{2007}]{suchanek2007yago}
Suchanek, F.~M.; Kasneci, G.; and Weikum, G.
\newblock 2007.
\newblock Yago: a core of semantic knowledge.
\newblock In {\em WWW},  697--706.
\newblock ACM.

\bibitem[\protect\citeauthoryear{Sun \bgroup et al\mbox.\egroup
  }{2019}]{sun2019rotate}
Sun, Z.; Deng, Z.-H.; Nie, J.-Y.; and Tang, J.
\newblock 2019.
\newblock Rotate: Knowledge graph embedding by relational rotation in complex
  space.
\newblock In {\em ICLR}.

\bibitem[\protect\citeauthoryear{Trouillon \bgroup et al\mbox.\egroup
  }{2016}]{trouillon2016complex}
Trouillon, T.; Welbl, J.; Riedel, S.; Gaussier, {\'E}.; and Bouchard, G.
\newblock 2016.
\newblock Complex embeddings for simple link prediction.
\newblock In {\em ICML},  2071--2080.

\bibitem[\protect\citeauthoryear{Vaswani \bgroup et al\mbox.\egroup
  }{2017}]{vaswani2017attention}
Vaswani, A.; Shazeer, N.; Parmar, N.; Uszkoreit, J.; Jones, L.; Gomez, A.~N.;
  Kaiser, {\L}.; and Polosukhin, I.
\newblock 2017.
\newblock Attention is all you need.
\newblock In {\em NIPS},  5998--6008.

\bibitem[\protect\citeauthoryear{Vig}{2019}]{vig2019transformervis}
Vig, J.
\newblock 2019.
\newblock A multiscale visualization of attention in the transformer model.
\newblock {\em arXiv preprint arXiv:1906.05714}.

\bibitem[\protect\citeauthoryear{Wang and Li}{2016}]{wang2016text}
Wang, Z., and Li, J.-Z.
\newblock 2016.
\newblock Text-enhanced representation learning for knowledge graph.
\newblock In {\em IJCAI},  1293--1299.

\bibitem[\protect\citeauthoryear{Wang \bgroup et al\mbox.\egroup
  }{2014a}]{wang2014knowledgeb}
Wang, Z.; Zhang, J.; Feng, J.; and Chen, Z.
\newblock 2014a.
\newblock Knowledge graph and text jointly embedding.
\newblock In {\em EMNLP}.

\bibitem[\protect\citeauthoryear{Wang \bgroup et al\mbox.\egroup
  }{2014b}]{wang2014knowledge}
Wang, Z.; Zhang, J.; Feng, J.; and Chen, Z.
\newblock 2014b.
\newblock Knowledge graph embedding by translating on hyperplanes.
\newblock In {\em AAAI}.

\bibitem[\protect\citeauthoryear{Wang \bgroup et al\mbox.\egroup
  }{2017}]{wang2017knowledge}
Wang, Q.; Mao, Z.; Wang, B.; and Guo, L.
\newblock 2017.
\newblock Knowledge graph embedding: A survey of approaches and applications.
\newblock {\em IEEE TKDE} 29(12):2724--2743.

\bibitem[\protect\citeauthoryear{Wang, Kulkarni, and
  Wang}{2018}]{wang2018dolores}
Wang, H.; Kulkarni, V.; and Wang, W.~Y.
\newblock 2018.
\newblock Dolores: Deep contextualized knowledge graph embeddings.
\newblock {\em arXiv preprint arXiv:1811.00147}.

\bibitem[\protect\citeauthoryear{Xiao \bgroup et al\mbox.\egroup
  }{2017}]{xiao2017ssp}
Xiao, H.; Huang, M.; Meng, L.; and Zhu, X.
\newblock 2017.
\newblock {SSP}: semantic space projection for knowledge graph embedding with
  text descriptions.
\newblock In {\em AAAI}.

\bibitem[\protect\citeauthoryear{Xiao, Huang, and Zhu}{2016}]{xiao2016transg}
Xiao, H.; Huang, M.; and Zhu, X.
\newblock 2016.
\newblock {TransG}: A generative model for knowledge graph embedding.
\newblock In {\em ACL}, volume~1,  2316--2325.

\bibitem[\protect\citeauthoryear{Xie \bgroup et al\mbox.\egroup
  }{2016}]{xie2016representation}
Xie, R.; Liu, Z.; Jia, J.; Luan, H.; and Sun, M.
\newblock 2016.
\newblock Representation learning of knowledge graphs with entity descriptions.
\newblock In {\em AAAI}.

\bibitem[\protect\citeauthoryear{Xie, Liu, and
  Sun}{2016}]{xie2016representationijcai}
Xie, R.; Liu, Z.; and Sun, M.
\newblock 2016.
\newblock Representation learning of knowledge graphs with hierarchical types.
\newblock In {\em IJCAI},  2965--2971.

\bibitem[\protect\citeauthoryear{Xu \bgroup et al\mbox.\egroup
  }{2017}]{xu2017knowledge}
Xu, J.; Qiu, X.; Chen, K.; and Huang, X.
\newblock 2017.
\newblock Knowledge graph representation with jointly structural and textual
  encoding.
\newblock In {\em IJCAI},  1318--1324.

\bibitem[\protect\citeauthoryear{Yang \bgroup et al\mbox.\egroup
  }{2015}]{yang2015embedding}
Yang, B.; Yih, W.-t.; He, X.; Gao, J.; and Deng, L.
\newblock 2015.
\newblock Embedding entities and relations for learning and inference in
  knowledge bases.
\newblock In {\em ICLR}.

\bibitem[\protect\citeauthoryear{Yang \bgroup et al\mbox.\egroup
  }{2019}]{yang2019xlnet}
Yang, Z.; Dai, Z.; Yang, Y.; Carbonell, J.; Salakhutdinov, R.; and Le, Q.~V.
\newblock 2019.
\newblock {XLNet}: Generalized autoregressive pretraining for language
  understanding.
\newblock {\em arXiv preprint arXiv:1906.08237}.

\bibitem[\protect\citeauthoryear{Zhang \bgroup et al\mbox.\egroup
  }{2016}]{zhang2016collaborative}
Zhang, F.; Yuan, N.~J.; Lian, D.; Xie, X.; and Ma, W.-Y.
\newblock 2016.
\newblock Collaborative knowledge base embedding for recommender systems.
\newblock In {\em KDD},  353--362.
\newblock ACM.

\bibitem[\protect\citeauthoryear{Zhang \bgroup et al\mbox.\egroup
  }{2018}]{zhang2018knowledge}
Zhang, Z.; Zhuang, F.; Qu, M.; Lin, F.; and He, Q.
\newblock 2018.
\newblock Knowledge graph embedding with hierarchical relation structure.
\newblock In {\em EMNLP},  3198--3207.

\bibitem[\protect\citeauthoryear{Zhang \bgroup et al\mbox.\egroup
  }{2019}]{zhang-etal-2019-ernie}
Zhang, Z.; Han, X.; Liu, Z.; Jiang, X.; Sun, M.; and Liu, Q.
\newblock 2019.
\newblock {ERNIE}: Enhanced language representation with informative entities.
\newblock In {\em ACL},  1441--1451.

\end{thebibliography}

\end{document}